\documentclass{article} 
\usepackage{iclr,times}
\usepackage{graphicx}

\usepackage{amsmath,amsfonts,bm}

\def\eqref#1{equation~\ref{#1}}

\def\1{\bm{1}}

\DeclareMathAlphabet{\mathsfit}{\encodingdefault}{\sfdefault}{m}{sl}
\SetMathAlphabet{\mathsfit}{bold}{\encodingdefault}{\sfdefault}{bx}{n}

\usepackage{hyperref}
\usepackage{url}
\usepackage{booktabs} 
\usepackage{multirow}
\usepackage{algorithm}
\usepackage{algpseudocode} 
\usepackage{tcolorbox}
\usepackage{xcolor}
\usepackage{booktabs}  
\usepackage{pifont}    
\usepackage{tabularx}  

\definecolor{CustomForestGreen}{RGB}{34,139,34} 
\definecolor{PromptBlue}{RGB}{58,110,175}

\title{LLMs Are Not Good Strategists, Yet Memory-Enhanced Agency Boosts Reasoning}

\author{Yi Wu \thanks{Equal contribution. The author order follows the alphabetical order of first names, and each author reserves the right to list their name first in their own records. It was published at the ICLR 2025 Workshop on Reasoning and Planning for LLMs.} \\
University of Chicago\\ 
\texttt{yiwu@uchicago.edu}\\
\And
Zhimin Hu \footnotemark[1]\\
University of Wisconsin-Madison \\
\texttt{hu436@wisc.edu}\\
}

\iclrfinalcopy 
\begin{document}

\maketitle

\begin{abstract}
Strategic reasoning in Large Language Models (LLMs) within long-horizon environments is often limited by inconsistent subgoals. In these settings, finite attention resources prevent the model from maintaining strategic coherence over thousands of steps. This limitation leads to strategic drift, where localized decisions fail to sustain a coherent trajectory across reasoning. To address this, we introduce EpicStar, a framework that enables agents to learn memory as policy to tackle long-horizon reasoning. Specifically, the agent maintains a bank of successful past episodes as a heuristic alongside a working memory to track short-term environmental changes. During inference, a dynamic gating mechanism determines whether to execute a retrieved action directly or to perform new reasoning through a contextual fusion of the retrieved episodes and current working memory. Utilizing StarCraft II as the testbed, we evaluated EpicStar against diverse opponent styles. It significantly outperforms baseline methods, achieving higher win rates while consuming an order of magnitude fewer tokens, and it maintains this advantage consistently across difficulty levels and opponent strategies. Our findings provide compelling evidence that structured cross-episode memory is essential for enabling LLM agents to perform robust, long-term strategic execution in dynamic, autonomous settings.
\end{abstract}

\section{Introduction}
Strategic reasoning in dynamic and partially observable environments represents a formidable challenge for Large Language Models (LLMs), where agents must maintain a coherent long-term trajectory while simultaneously adapting to rapid environmental shifts. StarCraft II has long served as a testbed for such capabilities because it demands the simultaneous coordination of resource management, technological expansion, and unit control over thousands of time steps. While LLMs have already achieved human-level reasoning in short-horizon settings \citep{mondorf2024comparing, kabra2025modeling}, their performance often degrades in long-range sequential reasoning tasks. We argue that the failures are mainly due to the fact that the agent tends to progressively overfit to local observations and consequently loses sight of its global objectives. 

Existing efforts to mitigate these issues either rely on more prompting loops to compress information \citep{llmsc2} or introduce more rules to stabilize the reasoning \citep{shao2024swarmbrain}. In this paper, we argue that robust strategic reasoning requires a transition from reactive prompting to structured reuse of past experience \citep{klein2017sources}. We introduce EpicStar, an agentic framework that integrates episodic retrieval with situational modulation as a policy for reasoning. EpicStar continuously stores successful gameplay episodes in a structured memory bank. During inference, the agent retrieves relevant episodes, acting as a strategic heuristic to guide its reasoning. To remain responsive to new situations, EpicStar maintains a working memory for tracking environmental changes and employs a dynamic gating mechanism to balance reusing past actions directly with new reasoning for situational adaptation. Additionally, in situational adaptation, retrieved episodes are fused via a context-aligned mechanism to provide high-level context and structure for ongoing reasoning.

\begin{figure*}[t]
\centering
\includegraphics[width=0.9\textwidth]{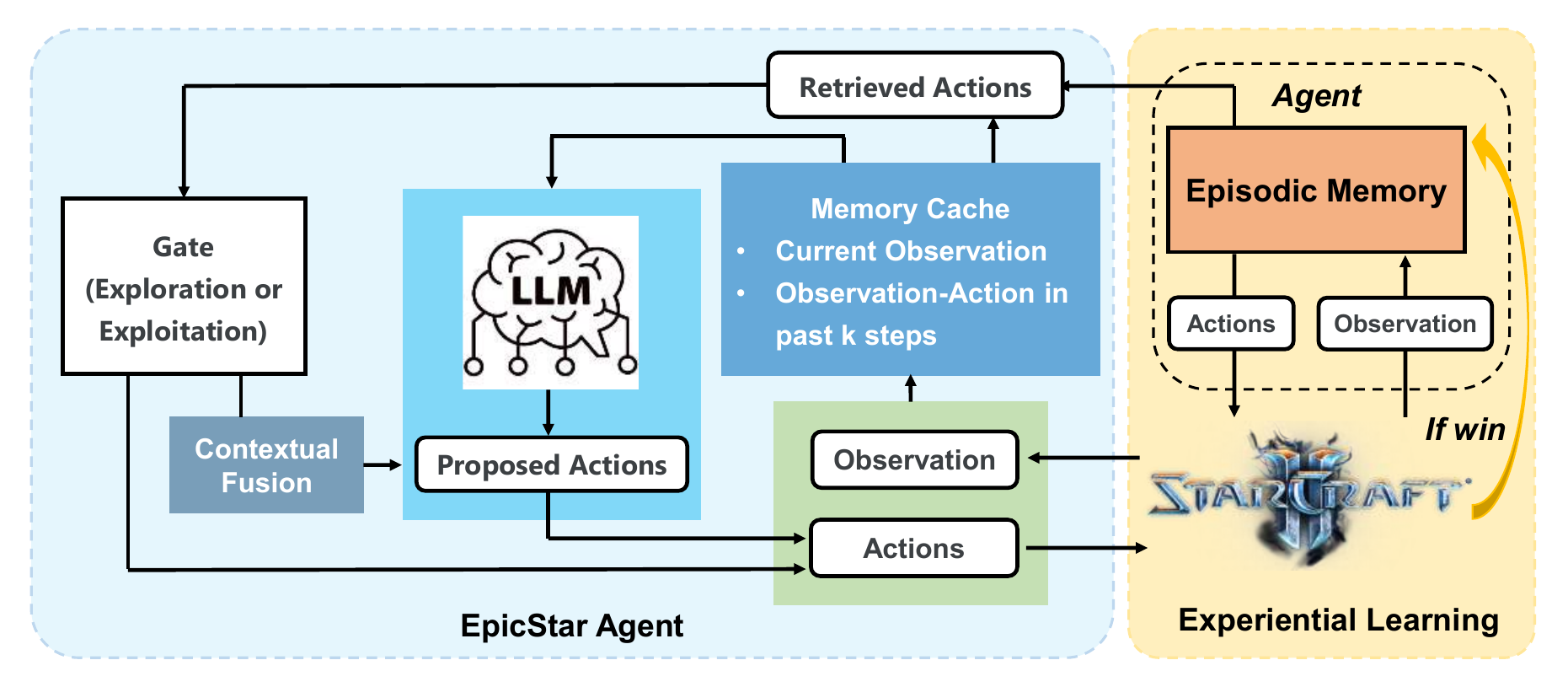}
    \caption{Overview of the EpicStar architecture. The agent is composed of distinct memory components and their interactions. Episodic memory stores gameplay episodes for retrieval and long-term learning. On the left blue panel, working memory consists of a short-term cache and gating mechanisms that manage exploration and exploitation, balancing coherent planning with dynamic adaptation. Contextual fusion enables bidirectional modulation between working memory and episodic memory, facilitating the integration of immediate context with past strategic knowledge.}
\label{tab:overview}
\end{figure*}

We evaluated EpicStar in StarCraft II using TextStarCraft II \citep{llmsc2} and chose its Chain of Summarization (CoS) method as our primary baseline, as it shares the most comparable assumptions and design scope with our method. We focus on level 5 and 6 game difficulties, which roughly correspond to entry-level and above-average human performance and mark the threshold at which CoS begins to show signs of performance degradation. We found that EpicStar consistently outperforms CoS across both levels, achieving higher win rates (Table~\ref{tab:agent_performance_comparison}) while consuming fewer tokens. Further, it demonstrates notably improved adaptability under diverse strategic conditions (Figure~\ref{fig:win_rate_details}).

In summary, our contributions are threefold: 1) we introduce EpicStar, an LLM-based agentic framework that incorporates structured episodic memory to support long-horizon strategic reasoning. EpicStar mitigates strategic drift by retrieving and adapting past trajectories to maintain coherence over time; 2) we propose a situational modulation mechanism that balances prior experience with real-time context. This includes a dynamic gating module that integrates episodic recall with working memory, and a contextual fusion step that translates retrieved episodes into strategic guidance; 3) we provide empirical evidence that both episodic and working memory are essential for strategic performance. Even a small number of high-quality episodic memories leads to marked improvements in win rate and adaptability. Code can be found at \href{https://github.com/ethanyiwu/EpicStar}{https://github.com/ethanyiwu/EpicStar}.

\section{Related Work}

\subsection{Game Agent}
The world of LLM-based game agents has been built on several fundamental ideas in agent design: 1) agents can learn from their action failures and reflect on their reasoning \citep{yao2022react,shinn2024reflexion}; agents can be equipped with external knowledge \citep{lewis2020retrieval}; agents can learn to use tools \citep{schick2023toolformer}. These advancements have led to the bloom of developing LLM agents in more complex game scenarios. To begin with, Voyager \citep{wang2023voyager} autonomously builds a toolkit of procedural skills through executable code in Minecraft. Conversely, PokeLLMon \citep{hu2024pok} retrieves external knowledge and uses memory for in-game reflection to refine high-level strategy in Pokémon. \citet{feng2023chessgpt} and \citet{huang2024pokergpt} utilize external gameplay datasets to train a language model to play Chess and Poker.

\subsection{StarCraft II Agent}
Among various games, StarCraft II has been unique in testing long-horizon strategic reasoning. As a real-time strategic game in partially observable environments, it requires players to reason under constraints from diverse sources of information and adapt strategies over extensive steps. The complexity of this domain was first made accessible to agentic research through the PySC2 interface \citep{pysc2}, which provides low-level state information and control. This interface fueled the development of reinforcement learning-based agents like AlphaStar \citep{Vinyals2019GrandmasterLI}, which achieved superhuman performance via massive-scale self-play, albeit at a prohibitive computational cost and with limited generalization. More recently, SwarmBrain \citep{shao2024swarmbrain} applied LLMs to this low-level interface. While it attempted to learn from gameplay through a log of semantic descriptions of strategies used, its learning is fundamentally oriented towards low-level control and lacks contextual richness.

Afterwards, the TextStarCraft II interface \citep{llmsc2} was introduced, moving away from micro-management and re-framing the game as a high-level strategic challenge. Our work is situated within this paradigm. Its CoS method serves as a strong baseline, using its sliding-window summary as a working memory to distill complex, high-dimensional state information into a concise, actionable representation. However, a clear gap remains: no existing agent for high-level strategic reasoning like StarCraft II learns from its own gameplay experience in a structured manner.

\section{Reasoning with Memory}
\subsection{Task Formulation}
Strategic reasoning within StarCraft II can be conceptualized as a sequential decision-making process characterized by the tuple $(S, A, T, R, Z, O)$. $S$ represents the actual state space of the world, where $S=\left\{s_1, s_2, \ldots, s_N\right\}$ is the set of all possible states the system can be in. $A$ represents the action space, and $A=\left\{a_1, a_2, \ldots, a_M\right\}$ is the set of actions the agent can take. In partially observable environments, the agent has access to an observation space $O=\left\{o_1, o_2, \ldots, o_K\right\}$. $T$ acts as the state transition function $T\left(s^{\prime} \mid s, a\right)=P\left(s^{\prime} \mid s, a\right)$ and $Z$ is the observation function, $Z\left(o \mid s^{\prime}, a\right)$, represents the probability of receiving observation 
$o$ after the agent takes action $a$ and transitions to state $s^{\prime}$. In this framework, the agent does not receive a direct reward $R(s,a)$ but rather observes the outcome of the game - either a win or a loss. Unlike traditional reinforcement learning, we do not directly estimate the $T$ and $Z$, instead, we rely on LLM reasoning coupled with memory mechanisms to maximize the expected accumulated reward.

\begin{algorithm}[!bp]
\caption{Working Memory Exploration and Exploitation - $W(\cdot)$}
\label{alg:working_memory}

\begin{algorithmic}
\Require Exploration Cool Down Frame $d_e$, Queue Pop Cool Down Frame $d_q$, Action Size $n$, Current Game Time $t$, Last Time to Explore $l_e$, Last Time to Pop Queue $l_q$
\end{algorithmic}
    
\begin{algorithmic}[1]  
    \State $(e_1, \dots, e_n) \leftarrow Re^{EC}(t,o_t)$ \Comment{Retrieve $n$ episodes}
    \State $a_t \leftarrow ExtractAction((e_1, \dots, e_n))$ \Comment{Select the first action}
    \If {$t \geq d_e + l_e$ }
        \State $l_e \leftarrow t$
        \State $o^{\prime}_t=o_t \times Q_{o}(t)$
        \State $(a_1, \dots, a_n) \leftarrow LLM(S, (  o^{\prime}_t))$ \Comment{Propose $n$ exploration actions}
        \State $Q_a.push((a_1, \dots, a_n))$ 
    \EndIf
    \If {$a_t =$ EmptyAction \textbf{and} $Q_a.size() > 0$ \textbf{and} $t \geq l_q + d_q$}
        \State $l_q = t$
        \State $a_t \leftarrow Q_a.pop()$ \Comment{Add the exploration action}
    \EndIf
    \State $t \leftarrow t + 1$
\State \textbf{return} $a_t$
\end{algorithmic}
\end{algorithm}

\subsection{Retrieval from Episodic Memory }
Given a set of gameplay data, we define a series of episodes $\{(t_i, o_i, a_i) \mid i = 1, 2, \ldots, e\}$, where $t_i$ represents the time in the game. The episodic memory, denoted as $M^{EC}(t, o, a)$, consists of episodes corresponding to victory games.

For each moment $(t, o)$ in a new game, we identify similar past moments by finding indices of episodes $D_{index}$ that closely match the current scenario from $M^{EC}$. Initially, we search $M^{EC}$ around $t$ to retrieve a subset:
\begin{equation}
{M^{EC}}_{t}=\operatorname{BinarySearch}(M^{EC}, t, t_\Delta)
\end{equation}
where $t_\Delta$ is a predefined parameter representing a time range around $t$ for the search (we set $t_\Delta=0$ for simplicity). Next, we compute the differences between the current observation $o_t$ and all previous observations $O' \in {M^{EC}}_t$, with each observation $o_t$ represented as a Python dictionary (with a unit as the key and a scalar value). We calculate two metrics: (1) the number of items that changed, denoted as $D_{item}$; and (2) the number of values that changed, denoted as $D_{value}$.
We then obtain the top $n$ memories based on the following criteria:
\begin{equation}
\begin{split}
D_{index}=\operatorname{Argsort}(\alpha \operatorname{MinMax}(D_{item}) \\
+ \beta \operatorname{MinMax}(D_{value}))
\end{split}
\end{equation}
where we perform min-max normalization on both $D_{item}$ and $D_{value}$, then sort $D$ in ascending order. We set $n = 3, \alpha = 0.5, \beta = 0.5$

Thus, the retrieval function is defined as:
\begin{equation}
    \begin{split}
        Re^{EC}(t, o) = 
        \begin{cases} 
        \Big\{\text{EmptyAction}\Big\},  \text{if }|{M^{EC}}_{t}|=0 \\
        \Big\{ {M^{EC}}_{t}(t_i, o_i, a_i) \mid i \in D_{index}[:n] \Big\},  \\\text{otherwise}
        \end{cases}
    \end{split}
\end{equation}

\subsection{Working Memory and Gating Mechanism}
For every time step $t$, the observation queue, $Q_{o}$, captures recent $k_{max}$ observations. Inspired by frame skipping \citep{frame_skipping}, the working memory recalls the past observations by a frame interval of $L$: 
\begin{equation}
Q_{o}(t)=\left\{\left(o_{t-k}\right) \mid k=L,2L, \ldots, k_{\max }L\right\}
\end{equation}
where $o_{t-k}$ represents observation at time $t-k$. Before reasoning, we map the observation $o_t$ to an augmented observation space defined as: $o^{\prime}_t=o_t \times Q_{o}(t)$. We set $k_{max} = 4$ and $L=24$.

We define the combination of the vanilla system prompt and contextual fusion as semantic knowledge $S$ and the querying of the LLM as the function $LLM(\cdot)$. We then engage in the exploration and exploitation process denoted as $W(\cdot)$ and initiate it with an exploration action queue, $Q_a$, as outlined in Algorithm~\ref{alg:working_memory}.

In $W(\cdot)$, $Q_o(t)$ provides short-term historical information that allows the agent to analyze the state’s tendency. In contrast, $Q_a$ stores actions for future interpolation into action sequences retrieved from episodic memory, enabling adaptive planning.

\begin{algorithm}[!bp]
\caption{EpicStar Agent}
\label{alg:training}

\begin{algorithmic}
\Require StarCraft II Game Environment $env$, Working Memory $W(\cdot)$
\end{algorithmic}
    
\begin{algorithmic}[1]  
    \State $env.initialize()$, 
    \State  $Q_a \leftarrow \emptyset$, $l_e \leftarrow 0$, $t \leftarrow 0$,  $l_q \leftarrow 0$ 
    \State $o_t \leftarrow env.observation()$ \Comment{Initial observation}

    \While {$env$ is not terminated}
        \State $a_t \leftarrow W(t, Q_a, o_t, Q_o(t), l_e, l_q)$
        \State $t \leftarrow t + 1$
        \State $o_t = env.step(a_t)$  \Comment{Next frame's observation}
    \EndWhile
\State \textbf{return} $env.game\_result()$
\end{algorithmic}
\end{algorithm}

\subsection{Contextual Fusion}
To make a fair comparison, we adapt the task instruction prompts from CoS methods \citep{llmsc2} as a backbone. Additionally, we incorporate contextual fusion to connect the information between working memory and episodic memory to the prompts. Specifically: 1) We augment the prompt with additional instructions that encourage the retrieved actions to be executable within the current scenario. This serves as a modulation from working memory to episodic memory. 2) To better align exploratory actions with episodic memory, we prompt the LLM to generate a high-level description of the strategy underlying the retrieved episodic memory and concatenate this information with the prompt. This serves as a modulation from episodic memory to working memory. The prompts are provided in the Appendix~\ref{appendix:prompt}. The full reasoning loop of our agent is presented in Algorithm~\ref{alg:training}.

\subsection{Learning Episodic Memory from Gameplay}
Episodic memory typically encompasses both learning and retrieval processes. To efficiently bootstrap the episodic memory, we use a rule-based agent to collect successful trajectories. Specifically, we gather game episodes as the agent competes against built-in opponents at Levels 6 and 7 for 20 rounds, and retain episodes from five winning games, yielding a total of 4,592 episodes. To ensure generalization and avoid overfitting to the data collection regime, we evaluate EpicStar on maps disjoint from those used during memory collection.

\subsection{Game Interface}
Given the action $a_t$ returned from Algorithm \ref{alg:training}, the game interface translates this action into specific procedures within StarCraft II. We adapt the game interface from \cite{llmsc2}, keeping the same action spaces for fair comparison. To improve generality, we removed overly specific operational routines hardcoded into the $attack$ and $defend$ actions, allowing them to be invoked more flexibly in different contexts.

\section{Experiments}

We evaluated EpicStar by comparing it against built-in agent opponents across varying difficulty levels, attack styles, and game maps in StarCraft II.

\subsection{Experimental Setup}
Our evaluation was carried out in two phases: (1) a comparative analysis against baseline methods and (2) an ablation study to assess the impact of individual components.

We used four closed-source OpenAI models spanning two generations, gpt-3.5-turbo, gpt-4-turbo, gpt-4o-mini, and gpt-4o \citep{hurst2024gpt}, to assess whether the benefits of EpicStar are consistent across backbones of varying capability. We conducted our experiments primarily at difficulty levels 5 and 6. To ensure consistency, we adopted the results from \cite{llmsc2} to mitigate potential performance degradation due to reproduction. Also, we only used \texttt{gpt-4-turbo} in Level 5 to match the results from CoS. Additional details on the experimental setup and metrics can be found in the Appendix~\ref{appendix:experiment_setup_metrics}.

\subsection{Evaluation Metrics}
We used the win rate as the primary indicator of agent performance. To further assess the efficiency of resource management and technological advancement within the game, we employed additional metrics detailed in Results\ref{results} and the Appendix~\ref{appendix:experiment_setup_metrics}.

\begin{table*}[!tb]
    \centering
    \footnotesize
    \renewcommand{\arraystretch}{1.2} 
    \begin{tabular}{ccccccc}
        \toprule
        Difficulty & Agent Type & Win Rate & PBR & RUR & APU & TR \\
        \midrule
        \multirow{7}{*}{Level 5}  & CoS(GPT-3.5-Turbo) & 0.550 & 0.0781 & 7875 & 0.7608 & 0.4476 \\
        & CoS(GPT-4-Turbo) & 0.600 & 0.0337 & 8306 & 0.7194 & 0.3452 \\
        \cline{2-7}
        & EpicStar(GPT-3.5-Turbo) & 0.575 & 0.1123 & 11363 & 0.7991 & 0.2500 \\
        & EpicStar(GPT-4-Turbo) & \textbf{0.750} & 0.1385 & 11483  & 0.8449 & 0.2524 \\
        & EpicStar(GPT-4o-mini) & 0.675 & 0.1211 & 9864 & 0.8123 & 0.2536 \\
        & EpicStar(GPT-4o) & 0.650 & 0.1175 & 11020 & 0.8107 & 0.2524 \\
        \midrule
        \multirow{5}{*}{Level 6} & CoS(GPT-3.5-Turbo) & 0.0833 & - & - & - & - \\
        \cline{2-7}
        & EpicStar(GPT-3.5-Turbo) & 0.150 & 0.1043 & 11191 & 0.7378 & 0.2375 \\
        & EpicStar(GPT-4o-mini) & \textbf{0.300} & 0.1089 & 10931 & 0.7865 & 0.2304 \\
        & EpicStar(GPT-4o) & 0.275 & 0.1176 & 10449 & 0.7404 & 0.2220 \\
        
        \bottomrule
    \end{tabular}
    \caption{Overall comparison of EpicStar with CoS baseline across different model backends against Level 5 and Level 6 built-in agents. Bold values indicate the highest win rates within each difficulty level.}
    \label{tab:agent_performance_comparison}
\end{table*}

\subsection{First Phase: Baseline Comparison}
In the first phase of our experiments, we compared EpicStar against the CoS baseline. The agent was evaluated against built-in agents at difficulty levels 5 and 6, with 40 evaluation rounds conducted at each level (except \texttt{gpt-4-turbo} is tested with 20 rounds due to its high cost). Specifically, we tested the agent against various built-in game strategies—\texttt{timing}, \texttt{rush}, \texttt{power}, \texttt{macro}, and \texttt{air}—as outlined in Appendix~\ref{appendix:experiment_setup_metrics}. These tests were conducted on two newly introduced maps for the 2024 Season, \texttt{Abyssal Reef LE} and \texttt{Ever Dream LE}. Each combination of strategy and map was tested four times, resulting in a total of 40 rounds.

\subsection{Second Phase: Ablation Study}
\label{ablation_study}
To evaluate the contribution of individual components within EpicStar, we selectively removed key elements while keeping other experimental conditions consistent with Phase 1 on \texttt{gpt-4o-mini}. The first ablation, denoted w/o exploration in the tables and figures below, disables the exploration step using information from working memory, so that the agent relies solely on retrieved episodes for reasoning. This allowed us to isolate and assess the role of exploration via LLMs within the framework. In a subsequent ablation, we removed the contextual fusion to examine the influence of bidirectional modulation between memory components on the agent’s reasoning and planning abilities. In EpicStar, ablating episodic memory reduces the agent to a configuration functionally similar to CoS. Therefore, we omit the ablation of the episodic memory component.

\section{Results}
\label{results}
\subsection{EpicStar v.s. Baseline}
We present a comprehensive analysis of EpicStar’s performance with CoS against the built-in agents, as shown in Table~\ref{tab:agent_performance_comparison}. At Level 5, EpicStar achieves a win rate of 67.5\% on \texttt{gpt-4o-mini} and 75.0\% on \texttt{gpt-4-turbo}, while at Level 6, it secures 30.0\% with \texttt{gpt-4o-mini}. In contrast, the strongest baseline, CoS (\texttt{gpt-4-turbo}), achieves a win rate of 60.0\% at Level 5. CoS, however, struggles considerably at Level 6, with a win rate of just 8.3\% (\texttt{gpt-3.5-Turbo}), and EpicStar nearly doubled the win rate to 15\% with the same model. Our method achieves a significant win rate gain compared to the baseline. Beyond the increase in win rate, our token consumption is only 14.5\% of CoS when using the same models (see more details in Appendix~\ref{appendix:comsuption})

Beyond the win rate and token consumption, we observe additional metrics that validate EpicStar’s enhanced performance. The Average Population Utilization (APU), which measures the efficiency of utilizing the population cap, reveals that EpicStar outperforms both baselines with values of 0.7991(\texttt{gpt-3.5-turbo}) and 0.8449(\texttt{gpt-4-turbo)}, compared to 0.7608 for CoS (\texttt{gpt-3.5-turbo}) and 0.7194 for CoS (\texttt{gpt-4-turbo}). A higher APU indicates more effective macro management to facilitate units in the game. We note occasional discrepancies among the Population Block Ratio (PBR; the lower the better), the Tech Rate (TR; a neutral indicator), and the Win Rate. These discrepancies occur because when EpicStar defeats the built-in agents, our agent does not accept their surrender. Thus, the game is extended until EpicStar reaches the population cap, which may result in lower PBR values even after a win.

\begin{table*}[!tb]
    \centering
    \footnotesize
    \renewcommand{\arraystretch}{1} 
    \begin{tabular}{ccccccc}
        \toprule
        Difficulty & Agent Type & Win Rate & PBR & RUR & APU & TR \\
        \midrule
        \multirow{3}{*}{Level 5}  & w/o exploration & 0.600 & 0.0598 & 11778 & 0.7628 & 0.2577 \\
        & w/o fusion  & 0.650 & 0.1292 & 11339 & 0.8179 & 0.2619 \\
        & EpicStar  & \textbf{0.675} & 0.1211 & 9864 & 0.8123 & 0.2536 \\
        \midrule
        \multirow{3}{*}{Level 6} & w/o exploration & 0.175 & 0.0522 & 10020 & 0.6828 & 0.2393 \\
        & w/o fusion  & 0.125 & 0.0992 & 13744 & 0.7497 & 0.2161 \\
        & EpicStar  & \textbf{0.300} & 0.1089 & 10931 & 0.7865 & 0.2304 \\
        \bottomrule
    \end{tabular}
    \caption{Overall comparison of EpicStar and two ablation agents against Level 5 and Level 6 built-in agents. We found that ablating either exploration using working memory or contextual fusion degrades the performance.}
    \label{tab:ablation_overview}
\end{table*}

\subsection{Ablation Study}
To understand the improved performance of EpicStar, we investigated the impacts of the exploration and contextual fusion components separately. As shown in Table \ref{tab:ablation_overview}, at Level 5, removing exploration decreases the win rate from 67.5\% (EpicStar) to 60.0\%, while removing contextual fusion slightly reduces it to 65.0\%. Thus, both components contribute positively, with exploration playing a slightly more critical role. At Level 6, this impact intensifies: the win rate significantly drops to 17.5\% without exploration and to 12.5\% without contextual fusion, compared to EpicStar's 30.0\%. These findings highlight that strategy coherence becomes increasingly important at higher difficulty levels, where misaligned exploration significantly impairs performance. It is also worth noting that other metrics are very close to each other; this is due to the similarities among the three agents.

\begin{figure*}[t]
  \centering
  \includegraphics[width=0.85\textwidth]{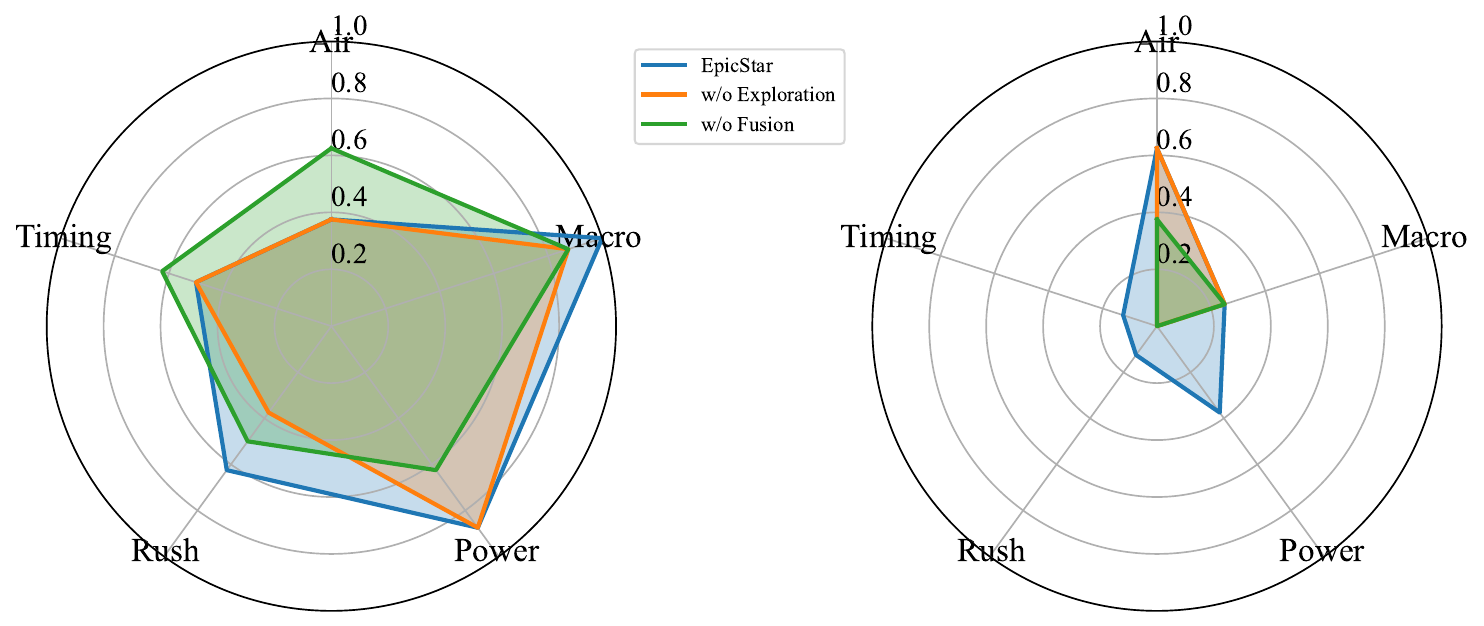}
  \caption{Detailed win rate across built-in strategies on EpicStar and two ablation agent types: agents without exploration and agents without contextual fusion.}
  \label{fig:win_rate_details}
\end{figure*}

Furthermore, as shown in Figure~\ref{fig:win_rate_details}, at Level 5 (left), the three agents exhibit comparable performance profiles across most strategies, indicating that the marginal benefit of exploration and contextual fusion is limited at this difficulty. At Level 6 (right), however, the gap widens substantially, particularly in the \texttt{Power}, \texttt{Rush}, and \texttt{Timing} strategies, where the ablated agents' win rates drop close to zero while EpicStar retains a modest but non-trivial win rate (e.g., 37.5\% in \texttt{Power}, 12.5\% in \texttt{Rush} and \texttt{Timing}). It is worth noting that absolute win rates remain relatively low for all agents at this difficulty, suggesting that Level 6 continues to pose a substantial challenge even for the full EpicStar agent; the results should therefore be interpreted as evidence of relative robustness rather than mastery of the task. This pattern suggests that the benefits of exploration and contextual fusion become more pronounced as task difficulty increases.

\subsection{Case Study}

\begin{figure*}[!tb]
    \centering
    \includegraphics[width=0.95\textwidth]{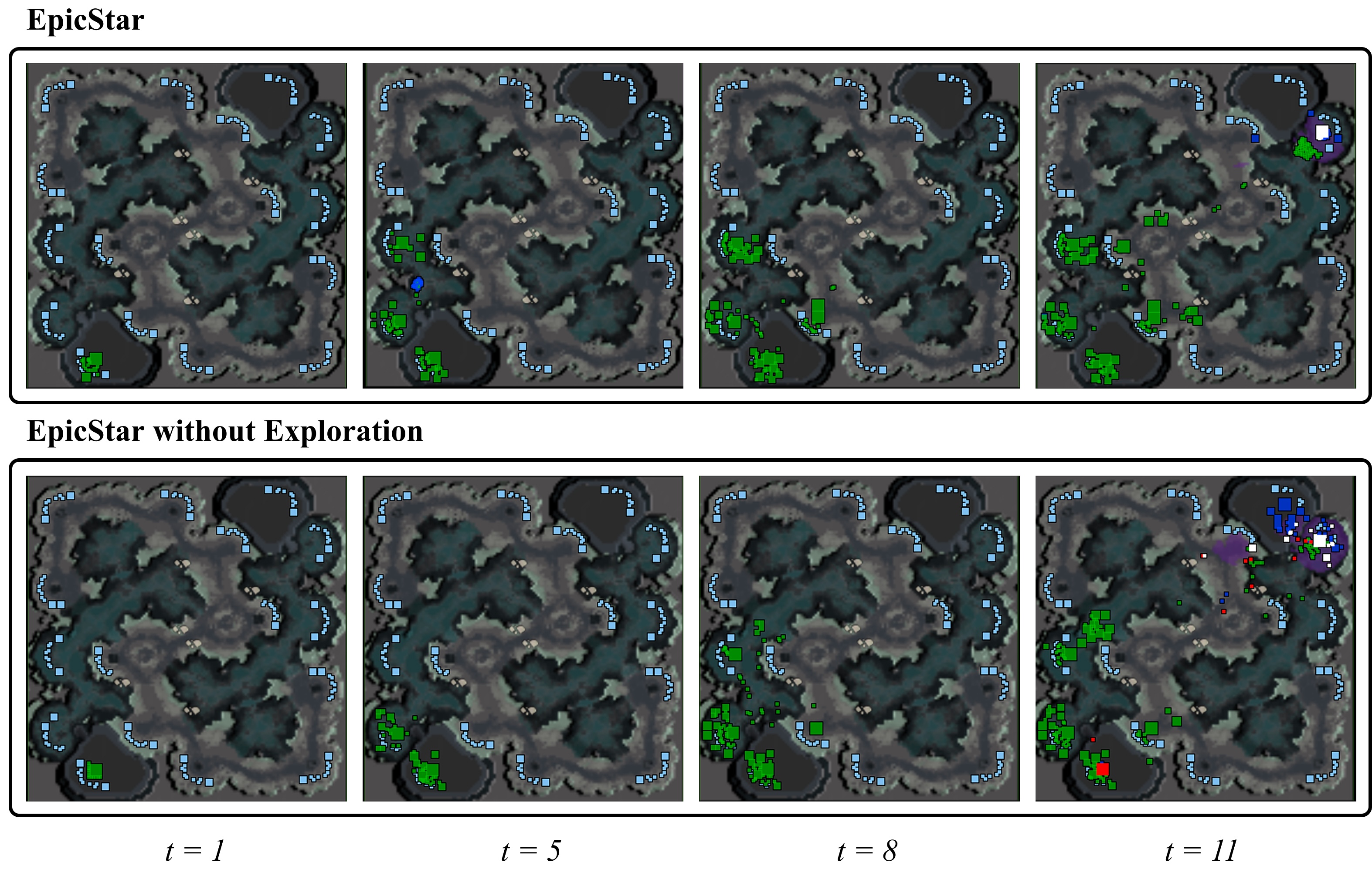}

    \caption{Expansion condition on $1,\ 5, 8, 11$ minutes for EpicStar (first row) and EpicStar without exploration (second row). The green dots represent the occupied bases, and EpicStar expands faster than EpicStar without exploration.}
    \label{fig:map_expansion}
\end{figure*}

Besides the quantitative results showing the improved performance of EpicStar compared to the baseline, we conducted a case study on the effect of working memory. Specifically, we compare EpicStar with and without exploration using working memory. We select a victory case against Level 6 air style built-in AI from each agent, and we find that EpicStar wins the game faster and achieves faster expansion, as illustrated(shown time 1:00, 5:00, 8:00, and 11:00, respectively), as shown in Figure \ref{fig:map_expansion}. This indicates that EpicStar reasons about its current strategic situation in the game and utilizes the limited resources more efficiently, and achieves better occupation, which leads to faster victory against the same opponent.

\section{Discussion \& Future Work}

Memory mechanisms have been explored in various short-horizon complex tasks. These studies primarily focus on utilizing memory as a storage for intermediate steps in multi-step reasoning \citep{wang2024symbolic} or as a means of grounding LLM inference with external knowledge \citep{wu2025airealtorgroundedpersuasive}. While \citet{hatalis2023memory} argue that LLM agents should possess long-term memory, the abundance of domain knowledge available online has led researchers to prioritize the efficient retrieval of existing knowledge over developing agents capable of autonomous learning and experiential integration. We argue that while enabling agents to learn from experience presents significant challenges, it is essential for the long-term advancement of autonomous intelligence. Our results suggest that even a small, curated bank of past trajectories can act as an effective substitute for costly exploration at inference time, echoing findings in case-based reasoning \citep{klein2017sources} that a handful of well-chosen precedents can carry more decision-relevant signal than exhaustive search. This reframes episodic memory not merely as a storage mechanism but as an implicit, non-parametric policy that complements the LLM's own reasoning.

However, our work is not without limitations. First, although we observed that incorporating exploration yields better performance than relying solely on episodic retrieval-a clear sign of improvement-there is currently limited empirical evidence to determine the extent to which this performance ceiling depends on the base model's capabilities. Meanwhile, our episodic memory was bootstrapped from a relatively small set of victories; we have not yet characterized how performance scales as the memory grows substantially larger, nor how the agent behaves when the bank contains noisy, redundant, or contradictory episodes.

Looking forward, several directions could extend this work. To enable more scalable memory use when reasoning, future work could investigate agentic frameworks that use specialized tokens to represent memory information, avoiding the growing prompt overhead of appending retrieved episodes in natural language as the memory bank scales. Additionally, one could explore methods to integrate parameterized memory in such experiential learning \citep{jun2025neural}, which remains a promising direction. Beyond StarCraft II, testing whether episodic-memory-as-policy transfers to other long-horizon, partially observable domains would help establish whether strategic drift and its mitigation via structured memory are general phenomena rather than artifacts of this particular testbed. Finally, an important safety consideration for any system that reuses past successful behavior is the risk of overfitting to opponent styles seen during memory collection; developing principled ways to detect when a retrieved episode no longer applies, rather than relying solely on the gating heuristic used here, would make such agents more robust when deployed against genuinely novel adversaries.

\section{Conclusion}
In this paper, we introduce a novel perspective on the design of reasoning agents for long-horizon tasks. By integrating episodic and working memory systems, we enable agents to maintain coherent strategic trajectories and adapt dynamically to evolving game scenarios. Empirical results demonstrate that our approach not only outperforms base methods in terms of performance but also has much lower token consumption, indicating that episodic memory can serve as an implicit mechanism for policy optimization, which enables more efficient reasoning. We show that even limited high-quality past experience can be maximized for significant performance gains when paired with a dynamic memory framework. This research also underscores the importance of human cognitive mechanisms in developing AI that can navigate and excel in complex strategic environments, bridging the gap between artificial intelligence and cognitive science.

\bibliography{epicstar}

@inproceedings{kabra2025modeling,
  title={Modeling Understanding of Story-Based Analogies Using Large Language Models},
  author={Kabra, Keshav and Inani, Kalit and Marupudi, Vijay and Varma, Sashank},
  booktitle={Proceedings of the Annual Meeting of the Cognitive Science Society},
  volume={47},
  year={2025}
}

@inproceedings{
llmsc2,
title={Large Language Models Play StarCraft {II}:Benchmarks and A Chain of Summarization Approach},
author={Weiyu Ma and Qirui Mi and Yongcheng Zeng and Xue Yan and Runji Lin and Yuqiao Wu and Jun Wang and Haifeng Zhang},
booktitle={The Thirty-eighth Annual Conference on Neural Information Processing Systems},
year={2024}
}

@article{Vinyals2019GrandmasterLI,
  title={Grandmaster level in StarCraft II using multi-agent reinforcement learning},
  author={Oriol Vinyals and Igor Babuschkin and Wojciech M. Czarnecki and Micha{\"e}l Mathieu and Andrew Dudzik and Junyoung Chung and David Choi and Richard Powell and others},
  journal={Nature},
  year={2019},
  volume={575},
  pages={350 - 354}
}

@article{hu2024pok,
  title={PokeLLMon: A Human-Parity Agent for Pokemon Battles with Large Language Models},
  author={Hu, Sihao and Huang, Tiansheng and Liu, Ling},
  journal={arXiv preprint arXiv:2402.01118},
  year={2024}
}

@article{shao2024swarmbrain,
  title={SwarmBrain: Embodied agent for real-time strategy game StarCraft II via large language models},
  author={Shao, Xiao and Jiang, Weifu and Zuo, Fei and Liu, Mengqing},
  journal={arXiv preprint arXiv:2401.17749},
  year={2024}
}

@inproceedings{frame_skipping,
  title={Frame Skip Is a Powerful Parameter for Learning to Play Atari},
  author={Alexander Braylan and Mark Hollenbeck and Elliot Meyerson and Risto Miikkulainen},
  booktitle={AAAI Workshop: Learning for General Competency in Video Games},
  year={2015}
}

@article{wang2023voyager,
  title={Voyager: An Open-Ended Embodied Agent with Large Language Models},
  author={Wang, Guanzhi and Xie, Yuqi and Jiang, Yunfan and Mandlekar, Ajay and Xiao, Chaowei and Zhu, Yuke and Fan, Linxi and Anandkumar, Anima},
  journal={Transactions on Machine Learning Research},
  year={2023}
}

@inproceedings{yao2022react,
  title={React: Synergizing reasoning and acting in language models},
  author={Yao, Shunyu and Zhao, Jeffrey and Yu, Dian and Du, Nan and Shafran, Izhak and Narasimhan, Karthik R and Cao, Yuan},
  booktitle={The eleventh international conference on learning representations},
  year={2022}
}

@article{shinn2024reflexion,
  title={Reflexion: Language agents with verbal reinforcement learning},
  author={Shinn, Noah and Cassano, Federico and Gopinath, Ashwin and Narasimhan, Karthik and Yao, Shunyu},
  journal={Advances in Neural Information Processing Systems},
  volume={36},
  year={2024}
}

@misc{pysc2,
  author       = {Google DeepMind},
  title        = {PySC2 - StarCraft II Learning Environment},
  year         = {2017},
  url          = {https://github.com/google-deepmind/pysc2},
  note         = {Accessed: 2024-12-27}
}

@article{schick2023toolformer,
  title={Toolformer: Language models can teach themselves to use tools},
  author={Schick, Timo and Dwivedi-Yu, Jane and Dess{\`\i}, Roberto and Raileanu, Roberta and Lomeli, Maria and Hambro, Eric and Zettlemoyer, Luke and Cancedda, Nicola and Scialom, Thomas},
  journal={Advances in neural information processing systems},
  volume={36},
  pages={68539--68551},
  year={2023}
}

@inproceedings{hatalis2023memory,
  title={Memory matters: The need to improve long-term memory in llm-agents},
  author={Hatalis, Kostas and Christou, Despina and Myers, Joshua and Jones, Steven and Lambert, Keith and Amos-Binks, Adam and Dannenhauer, Zohreh and Dannenhauer, Dustin},
  booktitle={Proceedings of the AAAI Symposium Series},
  volume={2},
  number={1},
  pages={277--280},
  year={2023}
}

@article{wu2025airealtorgroundedpersuasive,
  title={Ai realtor: Towards grounded persuasive language generation for automated copywriting},
  author={Wu, Jibang and Yang, Chenghao and Wu, Yi and Mahns, Simon and Wang, Chaoqi and Zhu, Hao and Fang, Fei and Xu, Haifeng},
  journal={arXiv preprint arXiv:2502.16810},
  year={2025}
}

@inproceedings{mondorf2024comparing,
  title={Comparing Inferential Strategies of Humans and Large Language Models in Deductive Reasoning},
  author={Mondorf, Philipp and Plank, Barbara},
  booktitle={Proceedings of the 62nd Annual Meeting of the Association for Computational Linguistics (Volume 1: Long Papers)},
  pages={9370--9402},
  year={2024}
}

@book{klein2017sources,
  title={Sources of power: How people make decisions},
  author={Klein, Gary A},
  year={2017},
  publisher={MIT press}
}

@article{lewis2020retrieval,
  title={Retrieval-augmented generation for knowledge-intensive nlp tasks},
  author={Lewis, Patrick and Perez, Ethan and Piktus, Aleksandra and Petroni, Fabio and Karpukhin, Vladimir and Goyal, Naman and K{\"u}ttler, Heinrich and Lewis, Mike and Yih, Wen-tau and Rockt{\"a}schel, Tim and others},
  journal={Advances in neural information processing systems},
  volume={33},
  pages={9459--9474},
  year={2020}
}

@article{feng2023chessgpt,
  title={Chessgpt: Bridging policy learning and language modeling},
  author={Feng, Xidong and Luo, Yicheng and Wang, Ziyan and Tang, Hongrui and Yang, Mengyue and Shao, Kun and Mguni, David and Du, Yali and Wang, Jun},
  journal={Advances in Neural Information Processing Systems},
  volume={36},
  pages={7216--7262},
  year={2023}
}

@article{huang2024pokergpt,
  title={PokerGPT: An End-to-End Lightweight Solver for Multi-Player Texas Hold'em via Large Language Model},
  author={Huang, Chenghao and Cao, Yanbo and Wen, Yinlong and Zhou, Tao and Zhang, Yanru},
  journal={arXiv preprint arXiv:2401.06781},
  year={2024}
}

@inproceedings{wang2024symbolic,
  title={Symbolic working memory enhances language models for complex rule application},
  author={Wang, Siyuan and Wei, Zhongyu and Choi, Yejin and Ren, Xiang},
  booktitle={Proceedings of the 2024 Conference on Empirical Methods in Natural Language Processing},
  pages={17583--17604},
  year={2024}
}

@inproceedings{jun2025neural,
  title={A Neural Network Model of Complementary Learning Systems: Pattern Separation and Completion for Continual Learning},
  author={Jun, James and Marupudi, Vijay and Shah, Raj Sanjay and Varma, Sashank},
  booktitle={Proceedings of the Annual Meeting of the Cognitive Science Society},
  volume={47},
  year={2025}
}

@article{hurst2024gpt,
  title={Gpt-4o system card},
  author={Hurst, Aaron and Lerer, Adam and Goucher, Adam P and Perelman, Adam and Ramesh, Aditya and Clark, Aidan and Ostrow, AJ and Welihinda, Akila and Hayes, Alan and Radford, Alec and others},
  journal={arXiv preprint arXiv:2410.21276},
  year={2024}
}
\bibliographystyle{iclr}

\newpage

\appendix

\section{Experimental Setup and Metrics Description}
\label{appendix:experiment_setup_metrics}

\subsection{Experimental Setup}

\textbf{Agent and Opponent Configuration:}  
To maintain a consistent and controlled testing environment, LLM agents are set to play as Protoss against built-in agent-controlled Zerg opponents. This arrangement enables a systematic evaluation of strategic performance across varying difficulty levels. The difficulty settings are listed below. 

\begin{table}[!htbp]
\centering
\footnotesize
\begin{tabular}{|c|c|c|}
\hline
\textbf{Level} & \textbf{BLZ Difficulty} & \textbf{Approximate League Equivalent} \\ \hline
1 & Very Easy & Bronze (Low) \\ \hline
2 & Easy & Bronze (Mid) \\ \hline
3 & Medium & Bronze (High) \\ \hline
4 & Hard & Silver (Low) \\ \hline
5 & Harder & Silver (Mid) \\ \hline
6 & Very Hard & Silver (High) \\ \hline
7 & Elite & Gold (Low) \\ \hline
8 & Cheat Vision & Gold (Mid) \\ \hline
9 & Cheat Money & Gold (High) \\ \hline
10 & Cheat Insane & Platinum (Low) \\ \hline
\end{tabular}
\caption{StarCraft II Built-in Agent Levels and Approximate League Equivalents}
\label{tab:ai_levels}
\end{table}

\textbf{Parameter Configuration:}  
The temperature parameter is set to 0.1 to prioritize strategic decision-making over random actions.

\textbf{Game Version:}  
All experiments were conducted using the latest Patch \texttt{5.0.14.93333} of StarCraft II.

\subsection{Evaluation Metrics}
\label{appendix:evaluation_metrics}  
Our evaluation framework for TextStarCraft II \cite{llmsc2} builds upon StarCraft II's established player performance analytics, incorporating tailored modifications to comprehensively assess LLM agent gameplay strategies.

\textbf{Win Rate:}  
This is the primary performance indicator for the agent in the game. It is calculated as the percentage of victories relative to the total games played.

\textbf{Population Block Ratio (PBR):}  
PBR assesses the agent's macro-management skills, specifically its ability to allocate resources efficiently and sustain population growth. It is defined as:

\begin{equation}
    \text{PBR} = \frac{\text{Time at Population Cap}}{\text{Game Duration}}
\end{equation}

This metric represents the proportion of time the agent spends at maximum population capacity relative to the total game duration until it reaches the 200/200 supply cap for the first time. A high PBR suggests ineffective macro-strategic planning and suboptimal decision-making.

\textbf{Resource Utilization Ratio (RUR):}  
RUR measures how efficiently the agent manages in-game resources over time. It is computed as:

\begin{equation}
    \text{RUR} = \frac{\text{Total Minerals + Total Gas Used}}{\text{Game Duration}}
\end{equation}

This metric evaluates the total resources expended relative to the game's duration until the agent first reaches the maximum supply. A high RUR may indicate poor resource utilization, reflecting suboptimal macro-strategic decisions.

\textbf{Average Population Utilization (APU):}  
APU quantifies how effectively the agent utilizes its available population capacity. It is calculated as:

\begin{equation}
    \text{APU} = \frac{1}{N} \sum_{i=1}^{N} \left( \frac{\text{Used Population at } i^{th} \text{ step}}{\text{Population Cap at } i^{th} \text{ step}} \right)
\end{equation}

This metric averages the ratio of the population used to total capacity across all time steps until the agent reaches full supply. A higher APU indicates more efficient population management and better macro-strategic execution.

\textbf{Technology Rate (TR):}  
TR evaluates the agent's inclination toward technological advancement by measuring its exploration of the technology tree. It is defined as:

\begin{equation}
    \text{TR} = \frac{\text{Completed Technologies}}{\text{Total Technologies Available}}
\end{equation}

This metric calculates the fraction of completed technologies and structures relative to the total available from start to finish of the game. TR reflects the agent's tendency to pursue technological upgrades, although it does not necessarily correlate with overall performance.

\subsection{StarCraft II Built-in Agent Styles}
\label{appendix:styles}
\begin{enumerate}
    \item \textbf{Timing}: Executes attacks at specific moments when a strategic advantage is perceived.
    \item \textbf{Rush}: Aims to overwhelm the opponent early by rapidly producing offensive units and launching swift attacks.
    \item \textbf{Power}: Focuses on building a strong economy and technological foundation before engaging in significant combat.
    \item \textbf{Macro}: Prioritizes long-term economic growth and infrastructure development.
    \item \textbf{Air}: Prioritizes the development and deployment of air units.
\end{enumerate}

\section{Prompt}
\label{appendix:prompt}
We present the prompt used in the EpicStar and CoS baseline \citep{llmsc2}. Prompt 1 (in Figure~\ref{fig:prompt1}) serves as the backbone prompt of EpicStar. In this prompt, the LLM will generate the strategy description based on the episodic memory retrieved. Here we present the example of the Warpgate strategy description generated by \texttt{gpt-4o}. Prompt 2 (in Figure~\ref{fig:prompt2}) is the CoS prompt and is used in our ablation study.

\begin{figure*}[t]
\begin{tcolorbox}[grow to right by=0cm, boxrule=0.6pt, colframe=PromptBlue!65, colback=PromptBlue!5, arc=2mm, left=6pt, right=6pt, top=6pt, bottom=6pt, fonttitle=\small\mdseries, title=Prompt 1: System prompt with contextual fusion from episodic memory (e.g. Warpgate strategy)., center title]
\tiny\ttfamily
You are an AI trained in analyzing and summarizing StarCraft II games. You understand the nuances and strategies of the Protoss race. \textbf{Specifically, you are playing the WarpGate Strategy in StarCraft II, which is a core tactic for the Protoss race, leveraging their ability to instantly warp in units anywhere on the map using Pylons or Warp Prisms. By converting Gateways into Warp Gates, Protoss players can quickly reinforce their army or apply pressure without needing to rally units across the map. This strategy emphasizes map control, timing attacks, and flexibility, often paired with strong early-game units like Zealots or Stalkers. Proper use of Warp Gates can overwhelm opponents with rapid unit production and strategic positioning.}
\\

Based on the summaries of multiple rounds in a game, we want you to analyze the game progression in a structured way. Your analysis should include the following aspects:
\\

1. Game Overview: Provide a brief overview of the current situation based on all the rounds.
\\

2. Current Game Stage: Determine the stage of the game based on the information of all rounds. Is it the early game, mid-game, or late game?\\

3. Our Situation: Describe our current status in terms of:\\

    3.1 Units and Buildings: Analyze the state of our units and buildings.\\
    
    3.2 Economy: Evaluate our economic condition, including resource collection and usage.\\
    
    3.3 Technology: Describe the status of our technological research and what technologies we have unlocked so far. Analyze our technology tree, indicating the available and potential upgrades or units.\\
    
4. Our Strategy: Infer our potential strategy based on our current situation and the information of all rounds.\\

5. Enemy's Strategy: Infer the enemy's potential strategy, based on the available information.\\

6. Key Information: Highlight the most important aspects from all rounds that have significantly influenced the game.\\

For Protoss, keep an eye on Nexus's energy to Chrono Boost important structures, and keep an eye on training units and building pylons.\\

Based on the game situation and strategies used by both sides, provide specific suggestions for the following areas:\\

1. Our Strategy: Propose adjustments to our current strategy to counter the enemy's moves and capitalize on our strengths.\\

2. Units and Buildings: Offer ways to enhance our unit composition and improve our building layout, suited to the current stage of the game.\\

3. Economy: Recommend better practices for resource gathering and usage, in line with our strategic needs.\\

4. Technology: Suggest focused research paths to gain technological advantages, considering our current research status and technology tree.\\

5. Feasibility: Based on current resources like mineral, gas, buildings, supplies, workers, use your knowledge in StarCraft II to brainstorm 3 coarse decisions that can be successfully executed and think and explain why.\\

5. Decisions: Lastly, consider the current situation and the suggestions provided, make 3 actionable and specific decisions from the action dictionary\{TRAIN UNIT: \{0: TRAIN PROBE, 1: TRAIN ZEALOT, 2: TRAIN ADEPT, 3: TRAIN STALKER, 4: TRAIN SENTRY, 5: TRAIN HIGHTEMPLAR, 6: TRAIN DARKTEMPLAR, 7: TRAIN VOIDRAY, 8: TRAIN CARRIER, 9: TRAIN TEMPEST, 10: TRAIN ORACLE, 11: TRAIN PHOENIX, 12: TRAIN MOTHERSHIP, 13: TRAIN OBSERVER, 14: TRAIN IMMORTAL, 15: TRAIN WARPPRISM, 16: TRAIN COLOSSUS, 17: TRAIN DISRUPTOR, 18: MORPH ARCHON\}, BUILD STRUCTURE: \{19: BUILD PYLON, 20: BUILD ASSIMILATOR, 21: BUILD NEXUS, 22: BUILD GATEWAY, 23: BUILD CYBERNETICSCORE, 24: BUILD FORGE, 25: BUILD TWILIGHTCOUNCIL, 26: BUILD ROBOTICSFACILITY, 27: BUILD STARGATE, 28: BUILD TEMPLARARCHIVE, 29: BUILD DARKSHRINE, 30: BUILD ROBOTICSBAY, 31: BUILD FLEETBEACON, 32: BUILD PHOTONCANNON, 33: BUILD SHIELDBATTERY\}, RESEARCH TECHNIQUE: \{34: RESEARCH WARPGATERESEARCH, 35: RESEARCH PROTOSSAIRWEAPONSLEVEL1, 36: RESEARCH PROTOSSAIRWEAPONSLEVEL2, 37: RESEARCH PROTOSSAIRWEAPONSLEVEL3, 38: RESEARCH PROTOSSAIRARMORSLEVEL1, 39: RESEARCH PROTOSSAIRARMORSLEVEL2, 40: RESEARCH PROTOSSAIRARMORSLEVEL3, 41: RESEARCH ADEPTPIERCINGATTACK, 42: RESEARCH BLINKTECH, 43: RESEARCH CHARGE, 44: RESEARCH PROTOSSGROUNDWEAPONSLEVEL1, 45: RESEARCH PROTOSSGROUNDWEAPONSLEVEL2, 46: RESEARCH PROTOSSGROUNDWEAPONSLEVEL3, 47: RESEARCH PROTOSSGROUNDARMORSLEVEL1, 48: RESEARCH PROTOSSGROUNDARMORSLEVEL2, 49: RESEARCH PROTOSSGROUNDARMORSLEVEL3, 50: RESEARCH PROTOSSSHIELDSLEVEL1, 51: RESEARCH PROTOSSSHIELDSLEVEL2, 52: RESEARCH PROTOSSSHIELDSLEVEL3, 53: RESEARCH EXTENDEDTHERMALLANCE, 54: RESEARCH GRAVITICDRIVE, 55: RESEARCH OBSERVERGRAVITICBOOSTER, 56: RESEARCH PSISTORMTECH, 57: RESEARCH VOIDRAYSPEEDUPGRADE, 58: RESEARCH PHOENIXRANGEUPGRADE, 59: RESEARCH TEMPESTGROUNDATTACKUPGRADE\}, OTHER ACTION: \{60: SCOUTING PROBE, 61: SCOUTING OBSERVER, 62: SCOUTING ZEALOT, 63: SCOUTING PHOENIX, 64: MULTI-ATTACK, 65: MULTI-RETREAT, 66: CHRONOBOOST NEXUS, 67: CHRONOBOOST CYBERNETICSCORE, 68: CHRONOBOOST TWILIGHTCOUNCIL, 69: CHRONOBOOST STARGATE, 70: CHRONOBOOST FORGE, 71: EMPTY ACTION\}\}. This dictionary comprises four categories of actions: unit production, building construction, technology research, and other actions. Remember to align these decisions with the current stage and WarpGate strategy of the game, and avoid proposing actions that are not currently feasible, such as actions requiring more resources than we have now, actions that are not in correct condition, etc. Remember to outline each action with 	\textless ACTION NAME \textgreater, surrounded by \textless and \textgreater.
\end{tcolorbox}
\caption{System prompt with contextual fusion from episodes where Warpgate strategy was used. Description is generated using \texttt{gpt-4o} when summarizing the strategy used in retrieved episodic memory (in bold)}
\label{fig:prompt1}
\end{figure*}

\begin{figure*}[t]
\begin{tcolorbox}[grow to right by=0cm, boxrule=0.6pt, colframe=PromptBlue!65, colback=PromptBlue!5, arc=2mm, left=6pt, right=6pt, top=6pt, bottom=6pt, fonttitle=\small\mdseries, title=Prompt 2: System prompt of Chain of Summarization (CoS), center title]
\tiny\ttfamily
You are an AI trained in analyzing and summarizing StarCraft II games. You understand the nuances and strategies of the Protoss race.
\\

Based on the summaries of multiple rounds in a game, we want you to analyze the game progression in a structured way. Your analysis should include the following aspects:
\\

1. Game Overview: Provide a brief overview of the current situation based on all the rounds.
\\

2. Current Game Stage: Determine the stage of the game based on the information of all rounds. Is it the early game, mid-game, or late game?\\

3. Our Situation: Describe our current status in terms of:\\

    3.1 Units and Buildings: Analyze the state of our units and buildings.\\
    
    3.2 Economy: Evaluate our economic condition, including resource collection and usage.\\
    
    3.3 Technology: Describe the status of our technological research and what technologies we have unlocked so far. Analyze our technology tree, indicating the available and potential upgrades or units.\\
    
4. Our Strategy: Infer our potential strategy based on our current situation and the information of all rounds.\\

5. Enemy's Strategy: Infer the enemy's potential strategy, based on the available information.\\

6. Key Information: Highlight the most important aspects from all rounds that have significantly influenced the game.\\

For Protoss, keep an eye on Nexus's energy to Chrono Boost important structures, and keep an eye on training units and building pylons.\\

Based on the game situation and strategies used by both sides, provide specific suggestions for the following areas:\\

1. Our Strategy: Propose adjustments to our current strategy to counter the enemy's moves and capitalize on our strengths.\\

2. Units and Buildings: Offer ways to enhance our unit composition and improve our building layout, suited to the current stage of the game.\\

3. Economy: Recommend better practices for resource gathering and usage, in line with our strategic needs.\\

4. Technology: Suggest focused research paths to gain technological advantages, considering our current research status and technology tree.\\

5. Feasibility: Based on current resources like mineral, gas, buildings, supplies, workers, use your knowledge in StarCraft II to brainstorm 3 coarse decisions that can be successfully executed and think and explain why.\\

5. Decisions: Lastly, consider the current situation and the suggestions provided, make 3 actionable and specific decisions from the action dictionary\{TRAIN UNIT: \{0: TRAIN PROBE, 1: TRAIN ZEALOT, 2: TRAIN ADEPT, 3: TRAIN STALKER, 4: TRAIN SENTRY, 5: TRAIN HIGHTEMPLAR, 6: TRAIN DARKTEMPLAR, 7: TRAIN VOIDRAY, 8: TRAIN CARRIER, 9: TRAIN TEMPEST, 10: TRAIN ORACLE, 11: TRAIN PHOENIX, 12: TRAIN MOTHERSHIP, 13: TRAIN OBSERVER, 14: TRAIN IMMORTAL, 15: TRAIN WARPPRISM, 16: TRAIN COLOSSUS, 17: TRAIN DISRUPTOR, 18: MORPH ARCHON\}, BUILD STRUCTURE: \{19: BUILD PYLON, 20: BUILD ASSIMILATOR, 21: BUILD NEXUS, 22: BUILD GATEWAY, 23: BUILD CYBERNETICSCORE, 24: BUILD FORGE, 25: BUILD TWILIGHTCOUNCIL, 26: BUILD ROBOTICSFACILITY, 27: BUILD STARGATE, 28: BUILD TEMPLARARCHIVE, 29: BUILD DARKSHRINE, 30: BUILD ROBOTICSBAY, 31: BUILD FLEETBEACON, 32: BUILD PHOTONCANNON, 33: BUILD SHIELDBATTERY\}, RESEARCH TECHNIQUE: \{34: RESEARCH WARPGATERESEARCH, 35: RESEARCH PROTOSSAIRWEAPONSLEVEL1, 36: RESEARCH PROTOSSAIRWEAPONSLEVEL2, 37: RESEARCH PROTOSSAIRWEAPONSLEVEL3, 38: RESEARCH PROTOSSAIRARMORSLEVEL1, 39: RESEARCH PROTOSSAIRARMORSLEVEL2, 40: RESEARCH PROTOSSAIRARMORSLEVEL3, 41: RESEARCH ADEPTPIERCINGATTACK, 42: RESEARCH BLINKTECH, 43: RESEARCH CHARGE, 44: RESEARCH PROTOSSGROUNDWEAPONSLEVEL1, 45: RESEARCH PROTOSSGROUNDWEAPONSLEVEL2, 46: RESEARCH PROTOSSGROUNDWEAPONSLEVEL3, 47: RESEARCH PROTOSSGROUNDARMORSLEVEL1, 48: RESEARCH PROTOSSGROUNDARMORSLEVEL2, 49: RESEARCH PROTOSSGROUNDARMORSLEVEL3, 50: RESEARCH PROTOSSSHIELDSLEVEL1, 51: RESEARCH PROTOSSSHIELDSLEVEL2, 52: RESEARCH PROTOSSSHIELDSLEVEL3, 53: RESEARCH EXTENDEDTHERMALLANCE, 54: RESEARCH GRAVITICDRIVE, 55: RESEARCH OBSERVERGRAVITICBOOSTER, 56: RESEARCH PSISTORMTECH, 57: RESEARCH VOIDRAYSPEEDUPGRADE, 58: RESEARCH PHOENIXRANGEUPGRADE, 59: RESEARCH TEMPESTGROUNDATTACKUPGRADE\}, OTHER ACTION: \{60: SCOUTING PROBE, 61: SCOUTING OBSERVER, 62: SCOUTING ZEALOT, 63: SCOUTING PHOENIX, 64: MULTI-ATTACK, 65: MULTI-RETREAT, 66: CHRONOBOOST NEXUS, 67: CHRONOBOOST CYBERNETICSCORE, 68: CHRONOBOOST TWILIGHTCOUNCIL, 69: CHRONOBOOST STARGATE, 70: CHRONOBOOST FORGE, 71: EMPTY ACTION\}\}. This dictionary comprises four categories of actions: unit production, building construction, technology research, and other actions. Remember to align these decisions with the current stage and WarpGate strategy of the game, and avoid proposing actions that are not currently feasible, such as actions requiring more resources than we have now, actions that are not in correct condition, etc. Remember to outline each action with 	\textless ACTION NAME \textgreater, surrounded by \textless and \textgreater.
\end{tcolorbox}
\caption{System prompt of CoS, which is used in the ablation study.}
\label{fig:prompt2}
\end{figure*}

\section{Token Consumption of EpicStar and CoS}
\label{appendix:comsuption}
By checking the token usage data from the API (\texttt{gpt-4o-mini}):

On January $21, 2025$, we conducted a total of five test experiments using the CoS baseline, consuming $2,592,427$ tokens, with an average of $2,592,427/5 = 518485.4$ tokens per game.

On February $14, 2025$, we ran 37 experiments with EpicStar, utilizing $2,776,468$ tokens, with an average of $2,776,468/37 = 75039.7$ tokens per game.

From this data, it is evident that our token consumption for EpicStar is nearly an order of magnitude lower than that for the CoS baseline.

\end{document}